\documentclass[10pt,twocolumn,letterpaper]{article}

\usepackage[pagenumbers]{cvpr} 

\usepackage{algorithm,algpseudocode}
\usepackage{algorithmicx}

\usepackage[normalem]{ulem}
\usepackage{ifthen}
\newcommand{\pr}[1]{\texttt{#1}}

\usepackage{xcolor}

\renewcommand{\empty}{\varnothing}  

\newcommand{\bs}[1]{{\boldsymbol{#1}}}
\newcommand{\deps}{\bs{\epsilon}_\theta}
\newcommand{\depsh}{\bs{\hat{\epsilon}}_\theta}
\newcommand{\prompt}[1]{``\emph{#1}"}
\newcommand{\afterfigure}{\vspace{-1em}}

\usepackage{graphicx}
\usepackage{amsmath}
\usepackage{amssymb}
\usepackage{booktabs}

\usepackage{sidecap}

\usepackage[pagebackref,breaklinks,colorlinks]{hyperref}

\usepackage[capitalize]{cleveref}
\crefname{section}{Sec.}{Secs.}
\Crefname{section}{Section}{Sections}
\Crefname{table}{Table}{Tables}
\crefname{table}{Tab.}{Tabs.}

\begin{document}

\title{Plug-and-Play Diffusion Features for Text-Driven Image-to-Image Translation}
\author{Narek Tumanyan$^*$\qquad Michal Geyer$^*$\qquad Shai Bagon \qquad Tali Dekel \\
{\small Weizmann Institute of Science} \\
{\small *Indicates equal contribution.} \\
{\small Project webpage: \url{https://pnp-diffusion.github.io}}
}

\twocolumn[{
\renewcommand\twocolumn[1][]{#1}
\maketitle
\centering
\vspace*{-0.73cm}
\includegraphics[width=1\textwidth]{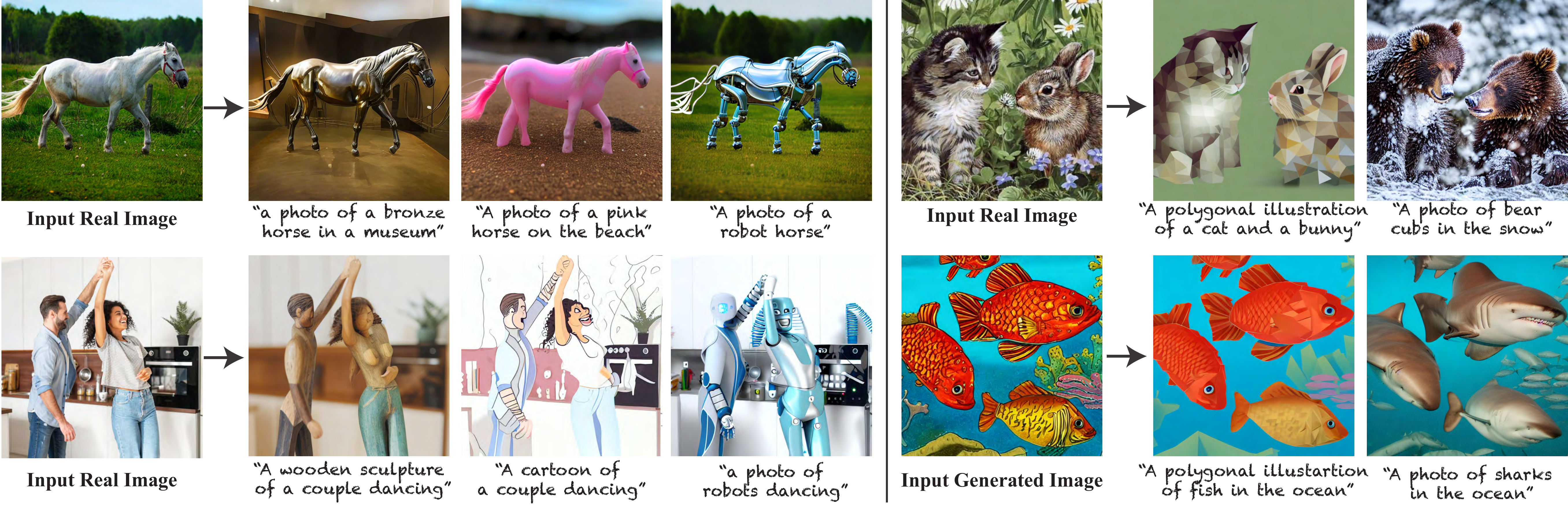} 
\captionof{figure}{Given a single real-world image as input,  our framework enables versatile text-guided translations of the original content. Our results exhibit high fidelity to the input structure and scene layout, while significantly changing the perceived semantic meaning of objects and their appearance.  Our method does not require any training, but rather harnesses the power of a pre-trained text-to-image diffusion model through its internal representation. We present new insights about deep features encoded in such models, and an effective framework to control the generation process through simple modification of these features. 
}\label{fig:teaser}
 \vspace*{0.27cm}
}]

\begin{abstract}
\vspace{-0.3cm}
   Large-scale text-to-image generative models have been a revolutionary breakthrough in the evolution of generative AI, allowing us to synthesize diverse images that convey highly complex visual concepts. However, a pivotal challenge in leveraging  such models for real-world content creation tasks is providing users with control over the generated content. In this paper, we present a new framework that takes text-to-image synthesis to the realm of image-to-image translation -- given a guidance image and a target text prompt as input, our method harnesses the power of a pre-trained text-to-image diffusion model to generate a new image that complies with the target text, while preserving the semantic layout of the guidance image. Specifically, we observe and empirically demonstrate that fine-grained control over the generated structure can be achieved by manipulating spatial features and their self-attention inside the model. This results in a simple and effective approach, where features extracted from the guidance image are directly injected into the generation process of the translated image, requiring no training or fine-tuning. We demonstrate high-quality results on versatile text-guided image translation tasks, including translating sketches, rough drawings and animations into realistic images,  changing of the class and appearance of objects in a given image, and modifications of global qualities such as lighting and color.
\end{abstract}

\vspace*{-3mm}
\section{Introduction}\label{sec:intro}
With the rise of text-to-image foundation models -- billion-parameter models trained on a massive amount of text-image data, it seems that we can translate our imagination into high-quality images through text \cite{ramesh2022dalle2,saharia202_imagen,gafni2022_make_a_scene,rombach2021ldm}. While such foundation models unlock a new world of creative processes in content creation, their power and expressivity come at the expense of user controllability, which is largely restricted to guiding the generation solely through an input text. In this paper, we focus on attaining control over the generated structure and semantic layout of the scene -- an imperative component in various real-world content creation tasks, ranging from  visual branding and marketing to digital art. That is, our goal is to take text-to-image generation to the realm of text-guided Image-to-Image (I2I) translation, where an input image guides the layout (e.g., the structure of the horse in Fig.~\ref{fig:teaser}), and the text guides the perceived semantics and appearance of the scene (e.g., ``robot horse'' in Fig.~\ref{fig:teaser}).

A possible approach for achieving control of the generated layout is to design text-to-image foundation models that explicitly incorporate additional guiding signals, such as user-provided masks~\cite{nichol2021glide,ramesh2022dalle2,gafni2022_make_a_scene}.
For example, recently Make-A-Scene~\cite{gafni2022_make_a_scene} trained a text-to-image model that is also conditioned on a label segmentation mask, defining the layout and the categories of objects in the scene.  However, such an approach requires an extensive compute as well as large-scale text-guidance-image training tuples, and can be applied at test-time to these specific types of inputs. In this paper, we are interested in a unified framework that can be applied to versatile I2I translation tasks, where the structure guidance signal ranges from artistic drawings to photorealistic images (see Fig.~\ref{fig:teaser}). Our method does not require any training or fine-tuning, but rather leverages a pre-trained and fixed text-to-image diffusion model~\cite{rombach2021ldm}.

Specifically, we pose the  fundamental question of how structure information is internally encoded in such a model. 
We dive into the intermediate spatial features that are formed during the generation process, empirically analyze them, and devise a new   framework that enables fine-grained control over the generated structure by applying simple manipulations to spatial features inside the model. Specifically, spatial features and their self-attentions are extracted from  the guidance image, and are directly injected into the text-guided generation process of the target image. We demonstrate that our approach is not only applicable in cases where the guidance image is generated from text, but also for real-world images that are inverted into the model.  

Our approach of operating in the space of diffusion features is related to Prompt-to-Prompt (P2P)~\cite{hertz2022prompt_to_prompt}, which recently observed that by manipulating the cross-attention layers, it is possible to control the relation between the spatial layout of the image to each word in the text.  We  demonstrate that fine-grained control over the generated layout is difficult to achieve solely from the interaction with a text. Intuitively, since the cross attention is formed by the association of spatial features to \emph{words}, it allows to capture rough regions at the \emph{object level},  yet localized spatial information that is not expressed in the source text prompt (e.g., object parts) is not guaranteed to be preserved by P2P. Instead, our method focuses only on \emph{spatial features} and their self-affinities -- we show that such features exhibit high granularity of spatial information, allowing us to control the generated structure, while not restricting the interaction with the text. Our method  outperforms P2P in terms of structure preservation and is superior in working with real guidance images.

\vspace*{1mm}\noindent{}To summarize, we make the following key contributions:\newline
(i)~We provide new empirical insights about internal spatial features formed during the diffusion process. \vspace{0.1cm}\newline
(ii)~We introduce an effective framework that leverages the power of pre-trained and fixed guided diffusion, allowing to perform high-quality text-guided I2I translation without any training or fine-tuning. \vspace{0.1cm}\newline
(iii)~We show, both quantitatively and qualitatively that our method outperforms existing state-of-the-art baselines, achieving significantly  better balance between preserving the guidance layout and deviating from its appearance.

\section{Related Work} \label{sec:related}

\paragraph{Image-to-image translation.} Image-to-Image (I2I) translation is aimed at estimating a mapping of an image from a source domain to a target domain, while preserving the domain-invariant characteristics of the input image, e.g., objects' structure or scene layout.  From classical  to  modern data-driven methods, numerous visual problems have been formulated and tackled as an I2I task (e.g., \cite{RaadG17,HertzmannJOCS01,ShihPDF13,ChenCTSH09,dekel2018contout2im}). Seminal deep-learning-based methods have proposed various GAN-based frameworks to  encourage the output image to comply with the  distribution of the target domain~\cite{zhu2017cyclegan,park2020cut,park2020swapping_ae,kim2022style}.
Nevertheless,  these methods require datasets of example images from both source and target domains, and often require training from scratch for each translation task (e.g., horse-to-zebra, day-to-night, summer-to-winter).  Other works utilize pre-trained 
GANs by performing the translation in its latent space~\cite{richardson2021encoding_in_style, tov2021_e4e,alaluf2021_restyle}.
Several methods have also considered the task of zero-shot I2I by training a generator on a single source-target image pair example~\cite{vinker2021_single_augmented,tumanyan2022splicing}.
With the advent of unconditional image diffusion models, several methods have been proposed to adopt or extend them for various I2I tasks \cite{saharia2022palette,wang2022piti}.
In this paper, we consider the task of \emph{text-guided image-to-image translation} where the target domain is not specified through a dataset of images but rather via a target text prompt. Our method is zero-shot, does not require training and is applicable to versatile I2I tasks.

\begin{figure*}[t!]
    \centering
    \includegraphics[width=\linewidth]{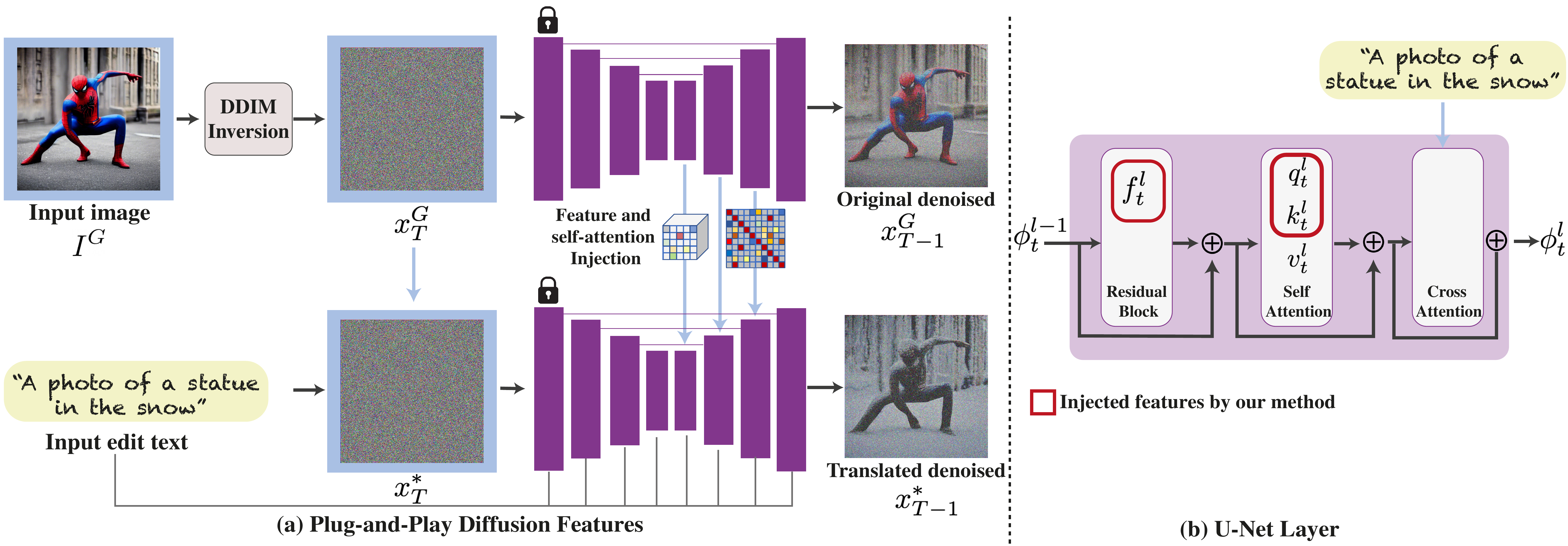}
    \caption{\emph{Plug-and-play Diffusion Features.} (a) Our framework takes  as input a guidance image and a text prompt describing the desired translation; the guidance image is inverted to initial noise $\bs{x}^G_T$, which is then progressively denoised using DDIM sampling. During this process, we extract $(\bs{f}_t^{l}, \bs{q}_t^{l}, \bs{k}_t^l)$ -- spatial features from the decoder layers and their self-attention, as illustrated in (b). To generate our text-guided translated image, we fix $\bs{x}^*_T=\bs{x}^G_T$ and inject the guidance features $(\bs{f}_t^{l}, \bs{q}_t^{l}, \bs{k}_t^l)$  at certain layers, as discussed in  Sec.~\ref{sec:method}. }
    \label{fig:pipeline}\afterfigure
\end{figure*}
\paragraph{Text-guided image manipulation.}
With the tremendous progress in language-vision models, a surge of methods have been proposed to perform various types of text-driven image edits. Various methods have proposed to combine CLIP~\cite{clip}, which provides a rich and powerful joint image-text embedding space, with a pre-trained unconditional image generator, e.g., a GAN~\cite{bau2021paint,StyleGanNada,StyleCLIP,liu2021fusedream} or a diffusion model~\cite{avrahami2022_blended_diff,kim2022diffusionclip,kwon2022diffusion_splice}. 
For example, DiffusionCLIP~\cite{kim2022diffusionclip} uses CLIP to fine-tune a diffusion model to perform text guided manipulations.
Concurrent to our work, \cite{kwon2022diffusion_splice} uses CLIP and semantic losses of~\cite{tumanyan2022splicing} to guide a diffusion process to perform I2I translation.
Aiming to edit the appearance of objects in real-world images, Text2LIVE~\cite{bar2022text2live} trains a generator on a single image-text pair, without additional training data; thus, avoiding the trade-off, inherent to pre-trained generators, between satisfying the target edit and maintaining high-fidelity to the original content.  While these methods have demonstrated impressive text-guided semantic edits, there is still a gap between the generative prior that is learned solely from visual data (typically on specific domains or ImageNet data), and the rich CLIP text-image guiding signal that has been learned from much broader and richer data.
Recently, text-to-image generative models have closed this gap by directly conditioning image generation on text during training \cite{nichol2021glide,ramesh2022dalle2,saharia202_imagen,gafni2022_make_a_scene,rombach2021ldm}.  These models have demonstrated unprecedented capabilities in generating high-quality and diverse images from text, capturing complex visual concepts (e.g., object interactions, geometry, or composition). Nevertheless, such models offer little control over the generated content. This creates a great interest in developing methods to adopt such unconstrained text-to-image models for controlled content creation.

Several \emph{concurrent} methods have taken first steps in this direction, aiming to influence different properties of the generated content~\cite{gal2022_textual_inversion,ruiz2022dreambooth,kawar2022imagic,wang2022piti}. DreamBooth~\cite{ruiz2022dreambooth} and Textual Inversion~\cite{gal2022_textual_inversion} share the same high-level goal of ``personalizing" a pre-trained text-to-image diffusion model given a few user-provided images. 
Our method also leverages a pre-trained text-to-image diffusion model to achieve our goal, yet does not involve any training or fine-tuning. Instead, we devise a simple framework that intervenes in the generation process by directly manipulating the spatial features.   

As discussed in \cref{sec:intro}, our methodological approach is related to Prompt-to-Prompt~\cite{hertz2022prompt_to_prompt}, yet our method offers several key advantages: (i)~enables fine-grained control over the generated shape and layout, (ii)~allows to use arbitrary text-prompts to express the target translation; in contrast to P2P that requires word-to-word alignment between a source and target text prompts, (iii)~demonstrates superior performance of real-world guidance images. 

Lastly, SDEdit~\cite{meng2021sdedit} is another method that applies edits on user provided images using free text prompts.  Their method noises the guidance image to an intermediate diffusion step, and then denoises it conditioned on the input prompt. 
This  simple approach leads to impressive results, yet exhibit a tradeoff between preserving the guidance layout and fulfilling the target text.   We demonstrate that our method  significantly outperforms SDEdit, providing better balance between these two ends.

\section{Preliminary}
\label{sec:prelimenary}

Diffusion models~\cite{sohl2015diffusion_basic,dhariwal2021diffusion_basic2,ho2020diffusion_basic3,rombach2021ldm} are probabilistic generative models in which an image is generated by progressively removing noise from an initial Gaussian noise image, $\bs{x}_T\!\sim\!\mathcal{N}(0, \mathbf{I})$.
These models are founded on two complementary random processes. 
the \emph{forward} process, in which Gaussian noise is progressively added to a clean image, $\bs{x}_0$:
\begin{equation}
    \bs{x}_t = \sqrt{\alpha_t}\cdot \bs{x}_{0} + \sqrt{1-\alpha_t}\cdot\bs{z} 
\end{equation} where $\bs{z}\!\sim\!\mathcal{N}(0, \mathbf{I})$ and $\left\{\alpha_t\right\}$ are the noise schedule.

The \emph{backward} process is aimed at gradually denoising $\bs{x}_T$, where at each step  a cleaner image is obtained. This process is achieved by a neural network $\deps(\bs{x}_t, t)$ that predicts the added noise $\bs{z}$.  Once trained, each step of the backward process consists of applying $\deps$ to the current $\bs{x}_t$, and adding a Gaussian noise perturbation 
to obtain a cleaner $\bs{x}_{t-1}$.

\begin{figure*}
\centering
\includegraphics[width=\linewidth]{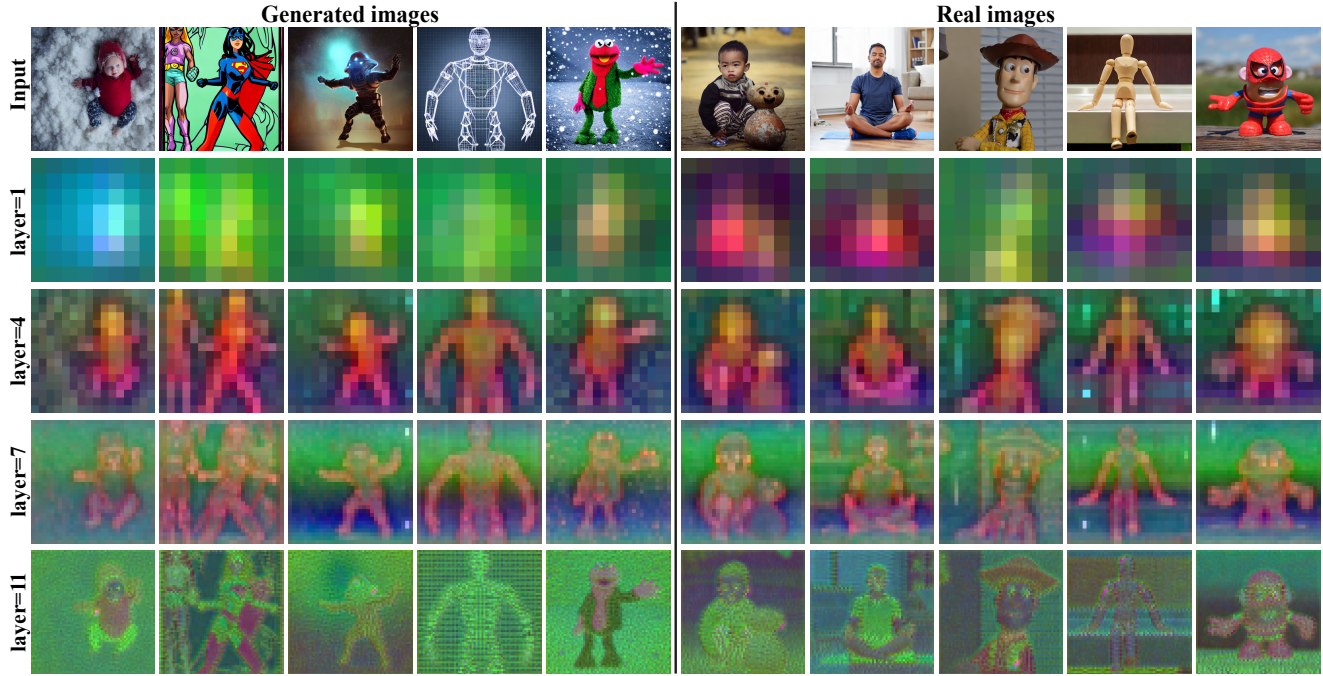} 
\caption{\emph{Visualising diffusion features.} We used a collection of 20 humanoid images (real and generated), and extracted spatial features from different decoder layers, at roughly 50\% of the generation process ($t=540$). For each block, we applied PCA on the extracted features across \emph{all} images and visualized the top three leading components. Intermediate features (layer 4)  reveal semantic regions (e.g., legs or torso) that are shared across all images, under large variations in  object appearance and the domain of images. Deeper features capture more high-frequency information, which eventually forms the output noise predicted by the model. See SM for additional visualizations. }
\label{fig:pca1}\afterfigure
\end{figure*}

Diffusion models are rapidly evolving and have been extended and trained to progressively generate images \emph{conditioned} on a guiding signal $\deps\!\left(\bs{x}_t,\bs{y}, t\right)$, e.g., conditioning the generation on another image~\cite{saharia2022palette}, class label~\cite{ho2022_class_cond_diff}, or text~\cite{nichol2021glide,ramesh2022dalle2,kim2022diffusionclip,rombach2021ldm}.

In this work, we leverage a pre-trained text-conditioned Latent Diffusion Model (LDM), a.k.a Stable Diffusion~\cite{rombach2021ldm}, in which the diffusion process is applied in the latent space of a pre-trained image autoencoder.   The model is based on a U-Net architecture~\cite{ronneberger2015unet} conditioned on the guiding prompt $P$. Layers of the U-Net comprise a residual block, a self-attention block, and a cross-attention block, as illustrated in \cref{fig:pipeline} (b).
The residual block convolve image features $\bs{\phi}^{l-1}_t$ from the previous layer $l\!-\!1$ to produce intermediate features $\bs{f}^l_t$. 
In the self-attention block,  features are projected into queries, $\bs{q}_t^l$, keys, $\bs{k}_t^l$, and values, $\bs{v}_t^l$, and the output of the block is given by:
\begin{equation}
    \bs{\hat{f}}^l_t = \bs{A}^l_t\bs{v}_t^l \mbox{\quad where\quad} \bs{A}^l_t = \mbox{Softmax}\!\left(\bs{q}_t^l {\bs{k}_t^l}^T \right)
    \label{eq:selfsim}
\end{equation} 
This operation allows for long-range interactions between image features.
Finally, cross-attention is computed between the spatial image features and the token embedding of the text prompt $P$.

\section{Method}
\label{sec:method}
Given an input guidance image $I^G$ and a target prompt $P$, our goal is to generate a new image $I^{*}$ that complies with $P$ and preserves the structure and semantic layout of $I^G$. 
We consider StableDiffusion~\cite{rombach2021ldm}, a state-of-the-art pre-trained and fixed text-to-image LDM model, denoted by $\deps\!\left(\bs{x}_t, P, t\right)$. 
This model is based on a U-Net architecture, as illustrated in  \cref{fig:pipeline} and discussed in \cref{sec:prelimenary}.

\begin{figure}[t!]
    \centering
    \includegraphics[width=\linewidth]{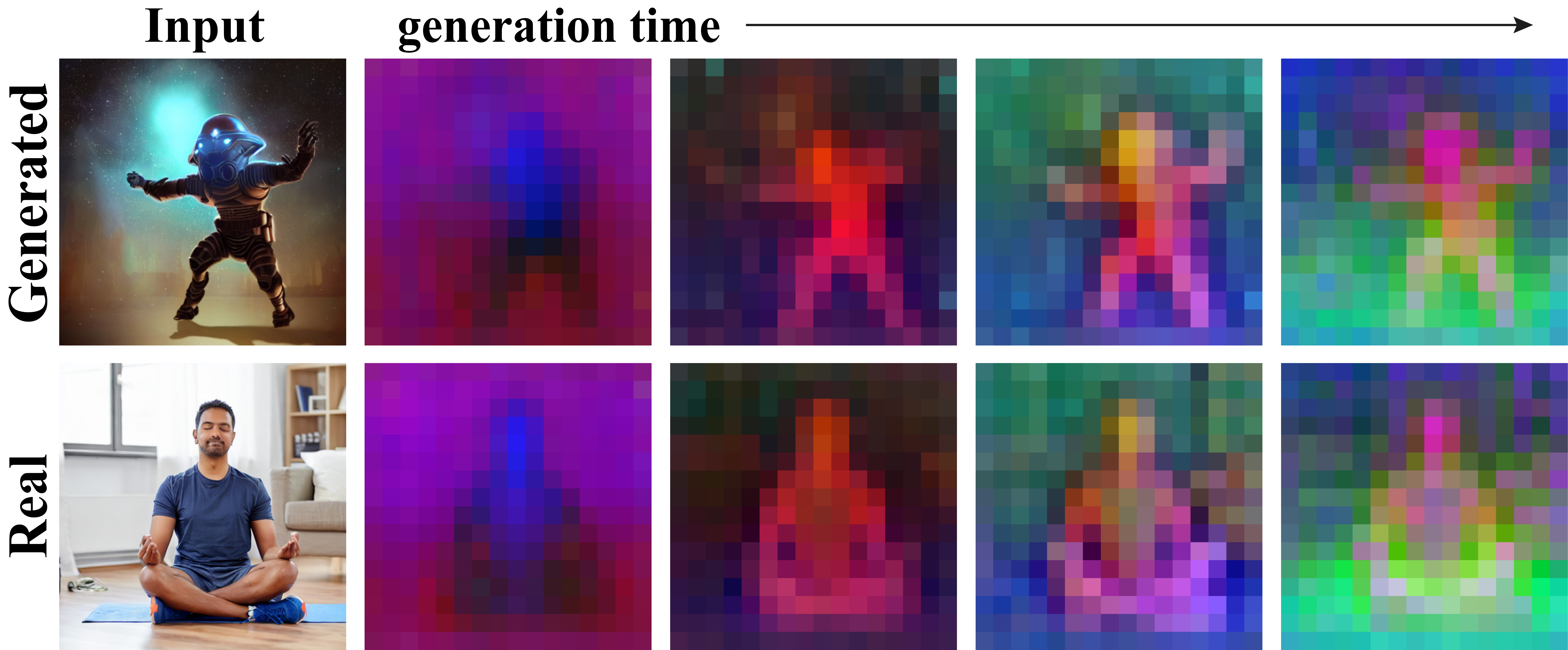}
    \caption{\emph{Diffusion features over generation time-steps.} Visualizing PCA of spatial features of layer $l\!=\!4$ for the humanoid images (\cref{fig:pca1}). Semantic parts are shared (have similar colors)  across images at each time step.}
    \label{fig:pca2} \afterfigure
\end{figure}
Our key finding is that fine-grained control over the generated structure can be achieved by manipulating spatial features inside the model during the generation process. Specifically, we observe and empirically demonstrate that: (i) spatial features extracted from intermediate decoder layers encode localized semantic information and are less affected by appearance information, and (ii) the self-attention, representing the affinities between the spatial features, allows to retain fine layout and shape details. 

Based on our findings, we devise a simple framework that extracts features from the  generation process of the guidance image $I^G$ and directly injects them along with $P$ into the generation process of $I^*$, requiring no training or fine-tuning (\cref{fig:pipeline}). Our approach is applicable for both text-generated and real-world guidance images, for which we apply DDIM inversion~\cite{song2020_ddim} to get the initial $\bs{x}^G_T$.

\begin{figure}[t!]
    \centering
    \includegraphics[width=\linewidth]{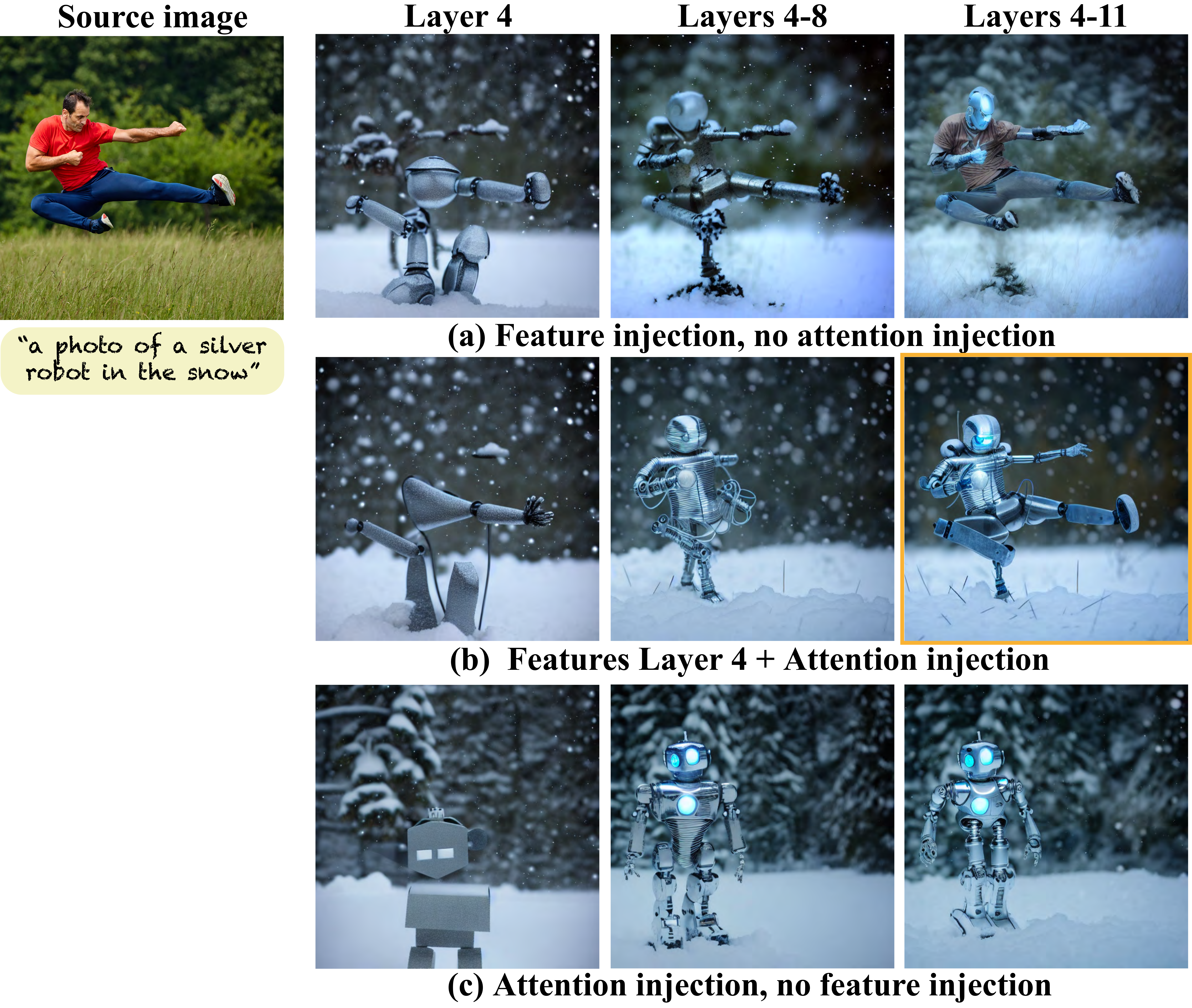}
    \caption{\emph{Ablating features and attention injection.}  (a) Features  extracted from the guidance image (left)  are injected into the generation process of the translated image (guided by a given text prompt).  While features at intermediate layers (\emph{Layer 4}) exhibit localized semantic information (Fig.~\ref{fig:pca1}), solely injecting these features is insufficient for retaining the guidance structure. Incorporating deeper (and higher resolution) features leads to better structure preservation, but results in appearance leakage from the guidance image to the generated one (\emph{Layers 4-11}). 
    (b) Injecting features only at layer 4 and self-attention maps at higher-resolution layers alleviates this issue. (c) Injecting only self-attention maps restricts the affinities between the features, yet there is no semantic association between the guidance features and the generated ones, resulting in misaligned structure. \emph{The result of our final configuration is highlighted in orange.} }
    \label{fig:ablation}\afterfigure
\end{figure}
\paragraph{Spatial features.}
In text-to-image generation, one can use descriptive text prompts to specify various scene and object proprieties, including those related to their shape, pose and scene layout, e.g.,  \prompt{a photo of a horse galloping in the forest}. However,  the exact scene layout, the shape of the object and its fine-grained pose often significantly vary across generated images from the same prompt under different initial noise $\bs{x}_T$. 
This suggests that the diffusion process itself and the resulting \emph{spatial} features  have a role in forming such fine-grained spatial information. This hypothesis is strengthened by~\cite{baranchuk2022labelefficient}, which demonstrated that semantic part segments can be estimated from  spatial features in an unconditional diffusion model.

We opt to gain a better understanding of how such semantic spatial information is internally encoded in $\deps$.  To this end, we perform a simple PCA analysis which allows us to reason about the visual properties dominating the high-dimensional features in $\deps$. Specifically, we generated a diverse set of images containing various humanoids in different styles, including both real and text-generated images; sample images are shown in \cref{fig:pca1}.
For each image, we extract features $\bs{f}_t^l$ from each layer of the decoder at each time step $t$, as illustrated in  \cref{fig:pipeline}(b). We then apply PCA on $\bs{f}_t^l$ across all images.

\cref{fig:pca1} shows the first three principal components for a representative subset of the images across different layers and a single time step.
As seen, the coarsest and shallowest layer is mostly dominated by foreground-background separation, depicting only a crude blob in the location of the foreground object. Interestingly, we can observe that the intermediate features (layer 4) encode localized semantic information shared across objects from different domains and  under significant appearance variations -- similar object parts (e.g., legs, torso, head) are depicted in similar colors \emph{across} all images (\emph{layer=4} row in \cref{fig:pca1}). These properties are consistent across the generation process as shown in \cref{fig:pca2}.
As we go deeper into the network, the features gradually capture more high-frequency low-level information which eventually forms the output noise  predicted by the network.  Extended feature visualizations can be found in the Supplementary Materials (SM) on our website.

\begin{figure}[t!]
    \centering    
    \includegraphics[width=\linewidth]{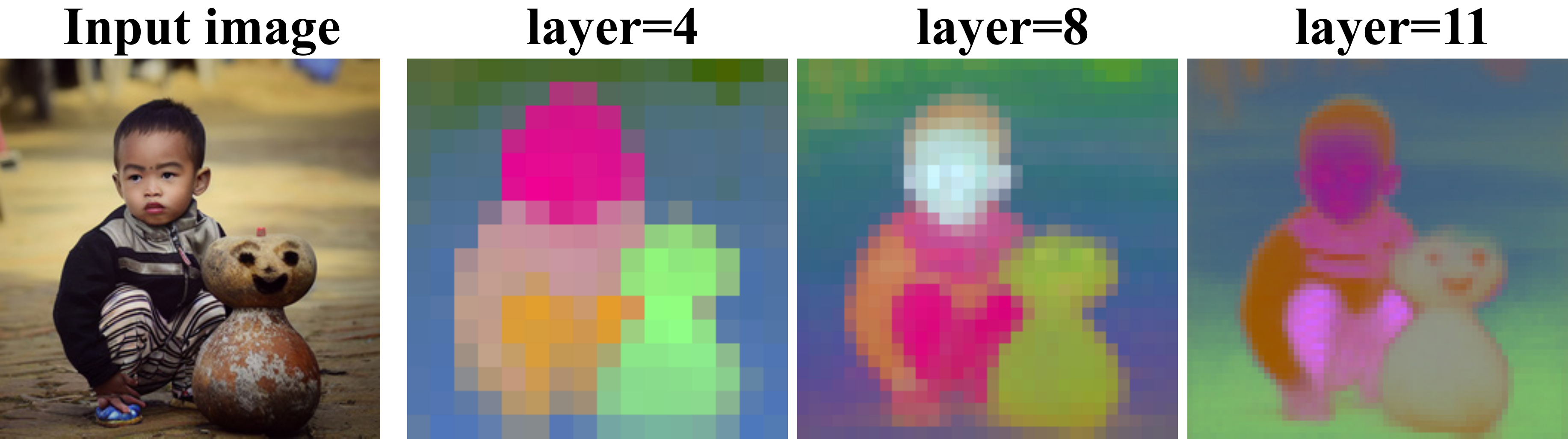}
    \caption{\emph{Self-attention visualization.} Showing 3 leading components of the self-attention matrix $\bs{A}^l_t$ computed for the input image for three different layers. The principal components are aligned with the layout of the image: similar regions share similar colors. Note how all pixels of the pants share similar color, despite their different appearance in the input image. }
    \label{fig:selfsim}\afterfigure
\end{figure}

\paragraph{Feature injection.} Based on these observations, we now turn to the translation task. Let $\bs{x}^G_T$ be the initial noise, obtained by  inverting $I^G$ using DDIM~\cite{song2020_ddim}.

Given the target prompt $P$, the generation of the translated image $I^*$ is carried with the same initial noise, i.e.,   $\bs{x}^*_T=\bs{x}^G_T$; we refer the reader to Appendix \ref{subsec:appendix-init-noise} for an analysis and justification of this design choice.

 At each step $t$ of the backward process, we extract the guidance features $\{\bs{f}_t^l\}$ from  the denoising step:  $\bs{z}_{t-1}^G=\deps(\bs{x}^G_t, \empty, t)$.\footnote{In the case of a \emph{generated} guidance image, $\bs{z}_{t-1}^G=\deps(\bs{x}^G_t, P_G, t)$, where $P_G$ is the text used to generate $I^G$. } These features  are then injected into the generation of $I^*$, i.e., in the denoising step of $\bs{x}^*_t$, we override the resulting features $\{\bs{f}^{*l}_t\}$ with $\{\bs{f}_t^l\}$. This operation is expressed by:
\begin{equation}
    \bs{z}^*_{t-1} = \bs{\hat{\epsilon}_\theta}(\bs{x}^*_t, P, t\;;\; \{\bs{f}_t^l\})
    \label{eq:injection}
\end{equation}
where we use $\bs{\hat{\epsilon}_\theta}(\cdot\,;\, \{\bs{f}_t^l\} )$ to denote the modified denoising step with the injected features $\{\bs{f}_t^l\}$.  In case of no injection, $\bs{\hat{\epsilon}_\theta}(\bs{x}_t, P, t\;;\; \empty)=\deps(\bs{x}_t, P, t)$.  

\cref{fig:ablation}(a) shows the effect of injecting spatial features $\bs{f}_t^l$ at increasing layers $l$.
As seen, injecting features only at layer $l\!=\!4$ is insufficient for preserving the structure of the guidance image. As we inject features in deeper layers,
the structure is better preserved, yet appearance information is leaked into the generated image (e.g., shades of the red t-shirt and blue jeans are apparent in  \emph{Layer 4-11}). To achieve a better balance between preserving the structure of $I^G$ and deviating from its appearance, we do not modify spatial features at deep layers, but  rather leverage the self-attention layers as discussed below.

\paragraph{Self-attention.}

Self-attention modules compute the \emph{affinities} $\bs{A}_t^l$ between the spatial features after linearly projecting them into queries and keys. These affinities have a tight connection to the established concept of self-similarly, which has been used to design structure descriptors by both classical and modern works \cite{shechtman2007selfsim,tumanyan2022splicing,kolkin2019strotss,bagon2010_sketch_the_common}. This motivates us to consider the attention matrices $\bs{A}^l_t$ to achieve fine-grained control over the generated content. 

\cref{fig:selfsim}, shows the leading principal components of a matrix $\bs{A}^l_t$ for a given image. As seen, 
in early layers, the attention is aligned with the semantic layout of the image, grouping regions according to semantic parts. Gradually, higher-frequency information is captured. 

Practically, injecting the self-attention matrix is done by replacing the matrix $\bs{A}^l_t$ in Eq.~\ref{eq:selfsim}. Intuitively, this operation  pulls features close together, according to the affinities encoded in $\bs{A}^l_t$. We denote this additional operation by modifying  \cref{eq:injection} as follows:
\begin{equation}
    \bs{z}^*_{t-1} = \bs{\hat{\epsilon}_\theta}(\bs{x}_t, P, t; \bs{f}_t^4,\{\bs{A}_t^l\})
    \label{eq:injection2}
\end{equation}
\cref{fig:ablation}(b) shows the effect of \cref{eq:injection2} for increasing injection layers;  
the maximal injection layer of $\bs{A}^l_t$ controls the level of fidelity to the original structure, while mitigating  the issue of appearance leakage. Fig.~\ref{fig:ablation}(c) demonstrates the pivotal role of  the features $\bs{f}_t^4$. As seen, with only self-attention, i.e.,  $\bs{z}^*_{t-1} = \bs{\hat{\epsilon}_\theta}(\bs{x}_t, P, t; \{\bs{A}_t^l\})$, there is no semantic association between the original content and the translated one, resulting in large deviations in structure. 

Our \emph{plug-and-play diffusion features} framework is summarized in Alg.~\ref{alg1}, and is controlled by two parameters: (i)~$\tau_f$ defines the sampling step $t$ until which $\bs{f}^4_t$ are injected. (ii)~$\tau_A$ is the sampling step until which $\bs{A}^l_t$ are injected. In all our results, we use a default setting where self-attention is injected into all the decoder layers. The exact settings of the parameters are discussed in \cref{sec:results}.

{\small \begin{algorithm}[t!]
\caption{Plug-and-Play Diffusion Features}\label{alg:pnp}
\begin{algorithmic}
\State \textbf{Inputs:} \\ $I^G$~~~$\triangleright$ real guidance image \\ 
$P$~~~$\triangleright$ target text prompt \\  $\tau_f$, $\tau_A$~~~$\triangleright$  injection thresholds\\
\State $\bs{x}_T^G \gets $ DDIM-inv($I^G$)
\State $\bs{x}^*_T \gets \bs{x}_T^G$ \Comment{Starting from same seed}
\For{$t \gets T\ldots 1$}
  \State $\bs{z}_{t-1}^G,\bs{f}^4_t,\left\{\bs{A}^l_t\right\} \gets \deps\!\left(\bs{x}_t^G,\empty,t\right)$
  \State $\bs{x}_{t-1}^G \gets \mbox{DDIM-samp}\!\left(\bs{x}_t^G,\bs{z}_{t-1}^G\right)$ 
  \State \textbf{if}~{$t > \tau_f$}~~\textbf{then}{~$\bs{f}^{*4}_t \gets \bs{f}^4_t$}~~\textbf{else}{~$\bs{f}^{*4}_t \gets \empty$}
  \State \textbf{if}~{$t > \tau_A$}~~\textbf{then}{~$\bs{A}^{*l}_t \gets \bs{A}^l_t$}~~\textbf{else}{~$\bs{A}^{*l}_t \gets \empty$}
  \State $\bs{z}^*_{t-1} \gets \depsh\!\left(\bs{x}^*_t,P,t\; ;\; \bs{f}^{*4}_t, \left\{\bs{A}^{*l}_t\right\}\right)$  
  \State $\bs{x}^*_{t-1} \gets \mbox{DDIM-samp}\!\left(\bs{x}^*_t,\bs{z}^*_{t-1}\right)$ 
\EndFor
\State \textbf{Output:} $I^* \gets \bs{x}^*_0$
\end{algorithmic}
\label{alg1}
\end{algorithm}}

\paragraph{Negative-prompting.}

In classifier-free guidance~\cite{ho2021classifier_free}, the predicted noise $\bs{\epsilon}$ at each sampling step is given by:
\begin{equation}
\bs{\epsilon}=w \deps(\bs{x}_t, P, t)+\left(1\!-\!w\right) \deps(\bs{x}_t, \empty,t)
\label{eq:class-free}
\end{equation}
where  $w\!>\!1$ is the guidance strength. 
That is, $\bs{\epsilon}$ is being extrapolated towards the conditional prediction $\deps(\bs{x}_t, P,t)$ and pushed away from the unconditional one $\deps(\bs{x}_t, \empty,t)$. This increases the fidelity of the denoised image to the prompt $P$, while allowing to deviate from  $\deps(\bs{x}_t, \empty,t)$.  Similarly, by replacing the empty prompt in \cref{eq:class-free} with a ``negative" prompt $P_n$, we can push away $\bs{\epsilon}$ from $\deps(\bs{x}_t, P_n,t)$. For example, using $P_n$ that describes the guidance image, we can steer the denoised image away from the original content. 

We use a parameter $\alpha\!\in\!\left[0,1\right]$ to balance between neutral and negative prompting:
\begin{equation}
\tilde{\bs{\epsilon}} =\alpha\deps\!\left(\bs{x}_t, \empty, t\right)+(1-\alpha)\deps\!\left(\bs{x}_t, P_n, t\right)
\label{eq:neg-prompt}
\end{equation}
We plug $\tilde{\bs{\epsilon}}$  instead of $\deps\!\left(\bs{x}_t, \empty,t\right)$ in \cref{eq:class-free}. That is, 
$\epsilon = w\deps\!\left(\bs{x}_t, P,t\right)+(1-w)\tilde{\bs{\epsilon}}$.

In practice, we find negative-prompting to be beneficial for handling textureless ``primitives'' guidance images (e.g., silhouette images). For natural-looking guidance images, it plays a minor role. See \cref{sec:appendix-np} for more details.

\section{Results}\label{sec:results}

We thoroughly evaluate our method both quantitatively and qualitatively on diverse guidance image domains, both real and generated ones, as discussed below. Please see \cref{sec:appendix-implementation} for full implementation details of our method. 

\paragraph{Datasets.} Our method supports versatile text-guided image-to-image translation tasks and can be applied to arbitrary image domains. Since there is no existing benchmark for such diverse settings, we created two new datasets: 
(i) \emph{Wild-TI2I}, comprises of 148 diverse text-image pairs, 53\% of which consists of real guidance images that we gathered from the Web;   (ii) \emph{ImageNet-R-TI2I}, a benchmark  we derived from the ImageNet-R dataset~\cite{hendrycks2021many}, which  comprises of various renditions (e.g., paintings, embroidery, etc.) of ImageNet object classes. To adopt this dataset for our purpose, we manually selected 3 high-quality images from 10 different classes. To generate our image-text examples, we created a list of text templates by defining for each source class target categories and styles, and automatically sampled their combinations. This results in total of 150 image-text pairs.
See Appendix \ref{sec:appendix-benchmarks} for full details. 

\begin{figure}[t!]
    \centering
    \includegraphics[width=\linewidth]{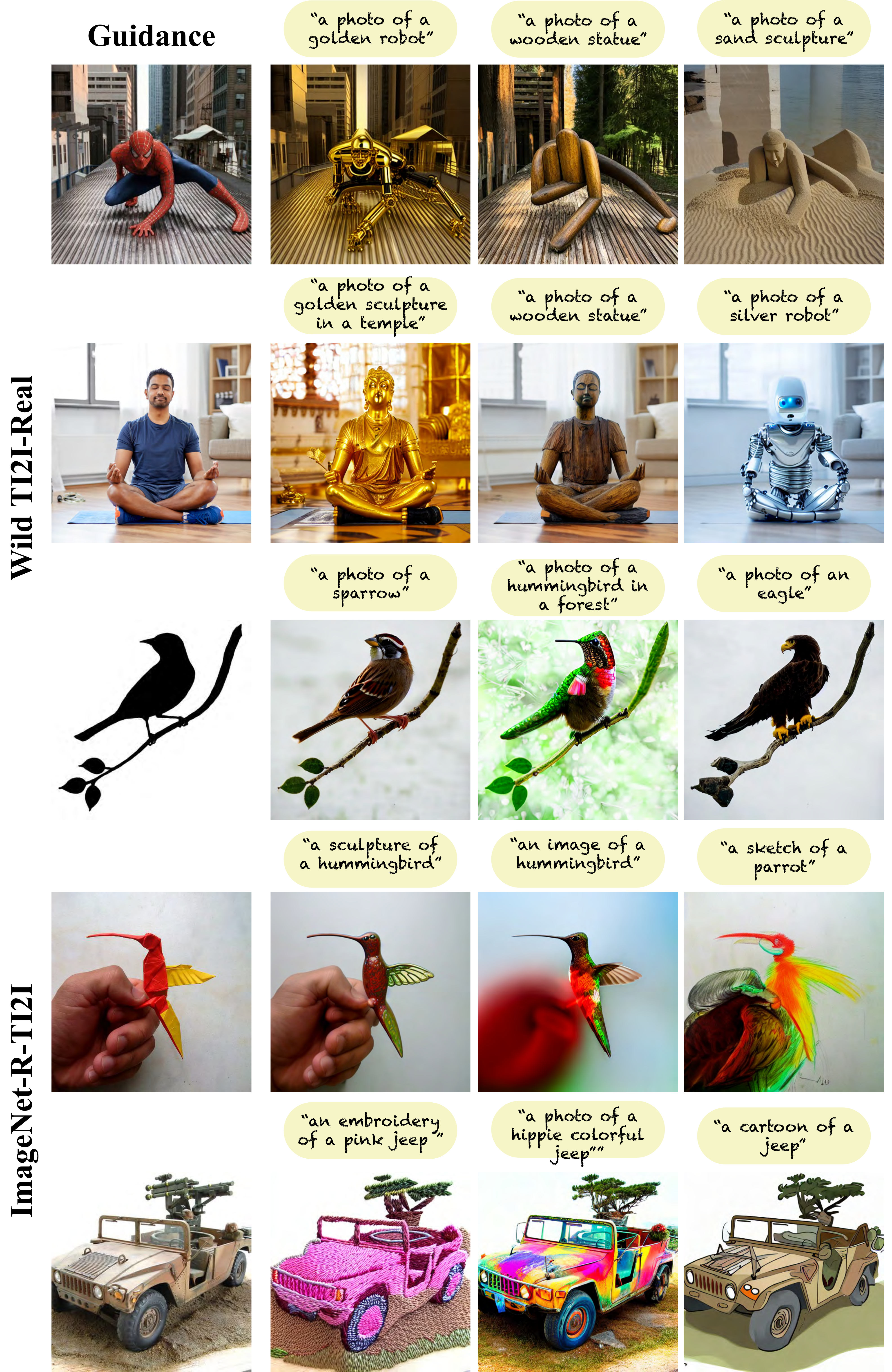}
    \caption{Sample results of our method on image-text pairs from \emph{Wild-TI2I} and \emph{ImageNet-R-TI2I} benchmarks.}
    \label{fig:results}\afterfigure
\end{figure}

\cref{fig:teaser,fig:results} show a sample of our results on both real and generated guidance images. 
Our results show both adherence to the guidance shape and compliance with different target prompts. Our method successfully handles both naturally looking as well as artistic and textureless guidance images.

\begin{figure*}[t!]
    \centering
    \includegraphics[width=\textwidth]{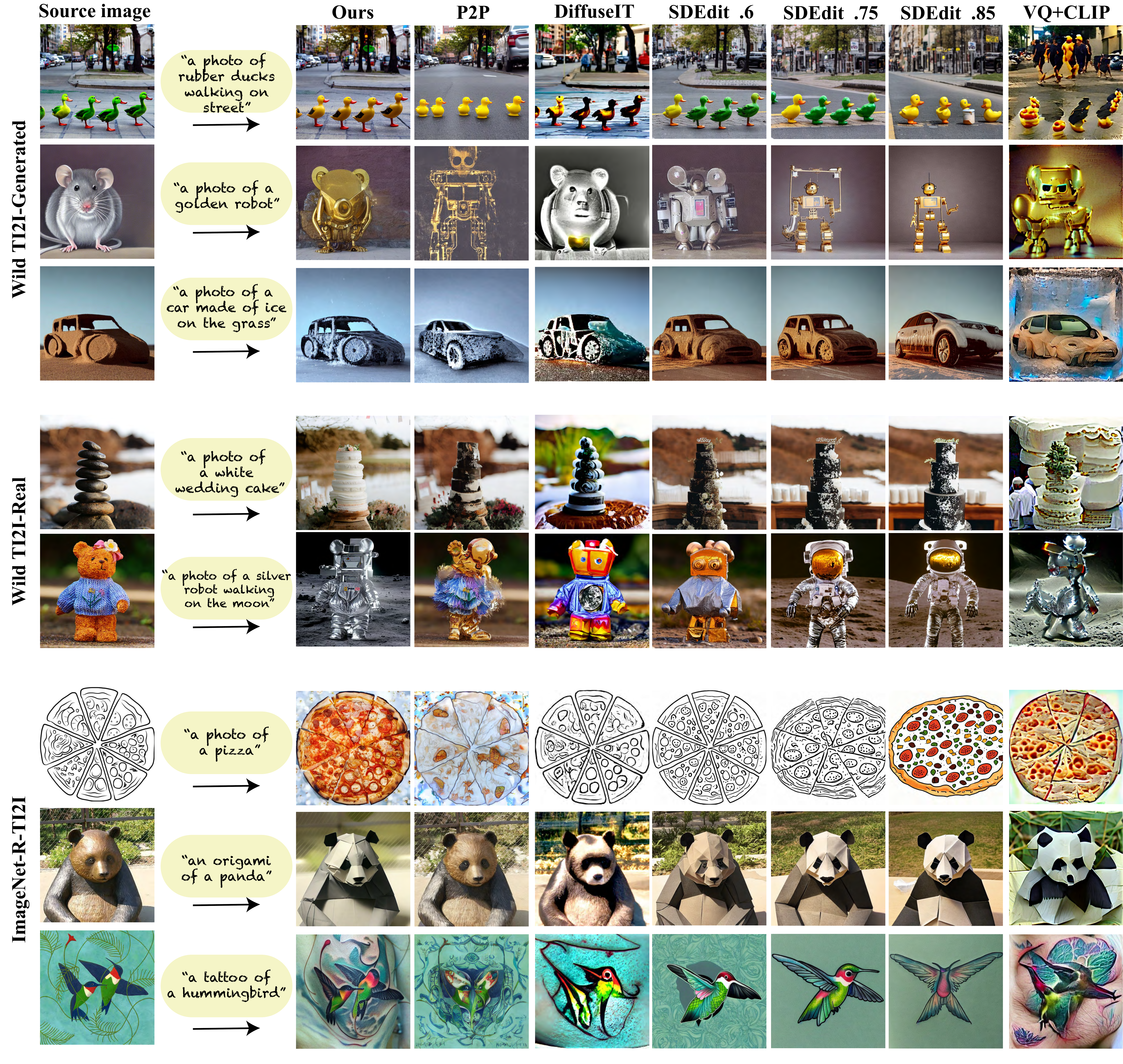} \vspace{-0.3cm}
    \caption{\emph{Comparisons.} Sample results are shown for each of the two benchmarks: \emph{ImageNet-R-TI2I} and \emph{Wild-TI2I}, which includes both real and generated guidance images. From left to right: the guidance image and text prompt, our results, P2P~\protect\cite{hertz2022prompt_to_prompt}, DiffuseIT~\protect\cite{kwon2022diffusion_splice},  SDedit~\protect\cite{meng2021sdedit} with 3 different noising levels, and VQ+CLIP~\protect\cite{crowson2022vqgan_clip}.}
    \label{fig:compare}\afterfigure
\end{figure*}

\subsection{Comparison to Prior/Concurrent Work}
We focus our comparisons on state-of-the-art baselines that can be applied to diverse text-guided I2I tasks, including: (i)~SDEdit~\cite{meng2021sdedit} under three different noising levels, (ii)~P2P~\cite{hertz2022prompt_to_prompt}, (iii)~DiffuseIT~\cite{kwon2022diffusion_splice}, and (iv)~VQGAN-CLIP~\cite{crowson2022vqgan_clip}.  We further provide qualitative comparisons to Text2LIVE~\cite{bar2022text2live}, FlexIT \cite{flexit2021} and DiffusionCLIP~\cite{kim2022diffusionclip}.  

We note that  P2P requires a source prompt that is word-aligned to the target prompt. Thus, we include a qualitative and quantitative comparison to P2P on our \emph{ImageNet-R-TI2I} benchmark, for which we automatically created aligned source-target prompts using the labels provided for the renditions and object categories. We further include qualitative comparison to  a subset of \emph{Wild-TI2I} for which the source and target prompts are aligned. For evaluating P2P on real guidance images, we applied DDIM inversion with the source text as in \cite{hertz2022prompt_to_prompt}.

Fig.~\ref{fig:compare} shows sample results of our method compared with the baselines. As seen, our method successfully translates diverse inputs, and works well for both real and generated guidance images. In all cases, our results exhibit both high preservation to the guidance layout and high fidelity to the target prompt. This is in contrast to SDEdit that suffers from an inherent tradeoff between the two -- with low noise level, the guidance structure is well preserved but in the expanse of hardly changing the  appearance; larger deviation in appearances can be achieved with higher noise level, yet the structure is damaged. VQGAN+CLIP exhibits the same behavior, with overall lower image quality. Similarly, DiffuseIT shows high fidelity to the guiding shape, with little changes to the appearance.

In comparison to P2P, it can be seen that their results on generated guidance images (first 3 rows) depict high fidelity to the target prompt, yet only rough preservation of layout, e.g., results in different number ducks (first row), or deviation from  the mouse shape (second row).   Furthermore, their method struggles to deviate from the guidance appearance and satisfy the target edit when it is applied to real images (4-8 rows). We  speculate that the reason is that DDIM inversion in their case is applied with a source text, requiring using low guidance scale at sampling. In contrast, our method performs DDIM inversion with an empty prompt, allowing us to use arbitrary guidance scale or prompts at generation.

We numerically evaluate these results using two complementary metrics: text-image CLIP cosine similarity to quantify how well the generated images comply with the text prompt (higher is better), and  distance between DINO-ViT self-similarity~\cite{tumanyan2022splicing}, to quantify the extent of structure preservation (lower is better). 

\begin{figure*}
    \centering
    \includegraphics[width=1\textwidth]{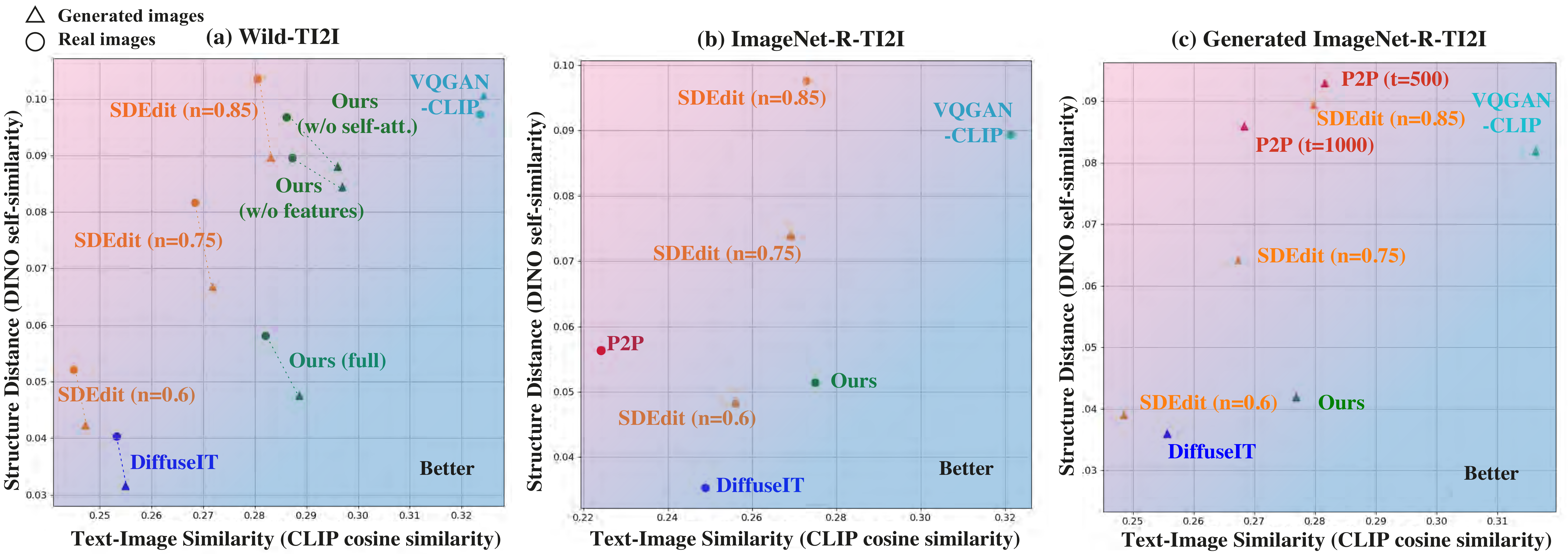}
   \caption{\emph{Quantitative evaluation}. We measure CLIP cosine similarity (higher is better) and DINO-ViT self-similarity distance (lower is better) to quantify the fidelity to text and preservation of structure, respectively. We report these metrics on three benchmarks: (a)  \emph{Wild-TI2I} for which  an ablation of our method is included, (b) \emph{ImageNet-R-TI2I}, and (c) \emph{Generated-ImageNet-R-TI2I}. Note that we could compare to P2P only for (b) and (c) due to their prompts restriction. All baselines struggle to achieve both low structure distance and a high CLIP score. Our method exhibit a better balance between these two ends across all benchmarks.
    }
    \label{fig:quant-comp}
   \afterfigure
\end{figure*}

As seen in \cref{fig:quant-comp}, our method outperforms the baselines by achieving both high preservation of structure (in par with SDEdit w/ very low noising level), and high fidelity to the target text (in par with SDEdit w/ very high noising level).  We note that VQGAN-CLIP and DiffuseIT  directly use the evaluation metrics as their objective (CLIP loss in~\cite{crowson2022vqgan_clip} and DINO self-similarity in~\cite{kwon2022diffusion_splice}), which explains their respective scores in these metrics. 

\begin{figure}[t!]
    \centering
    \includegraphics[width=.9\linewidth]{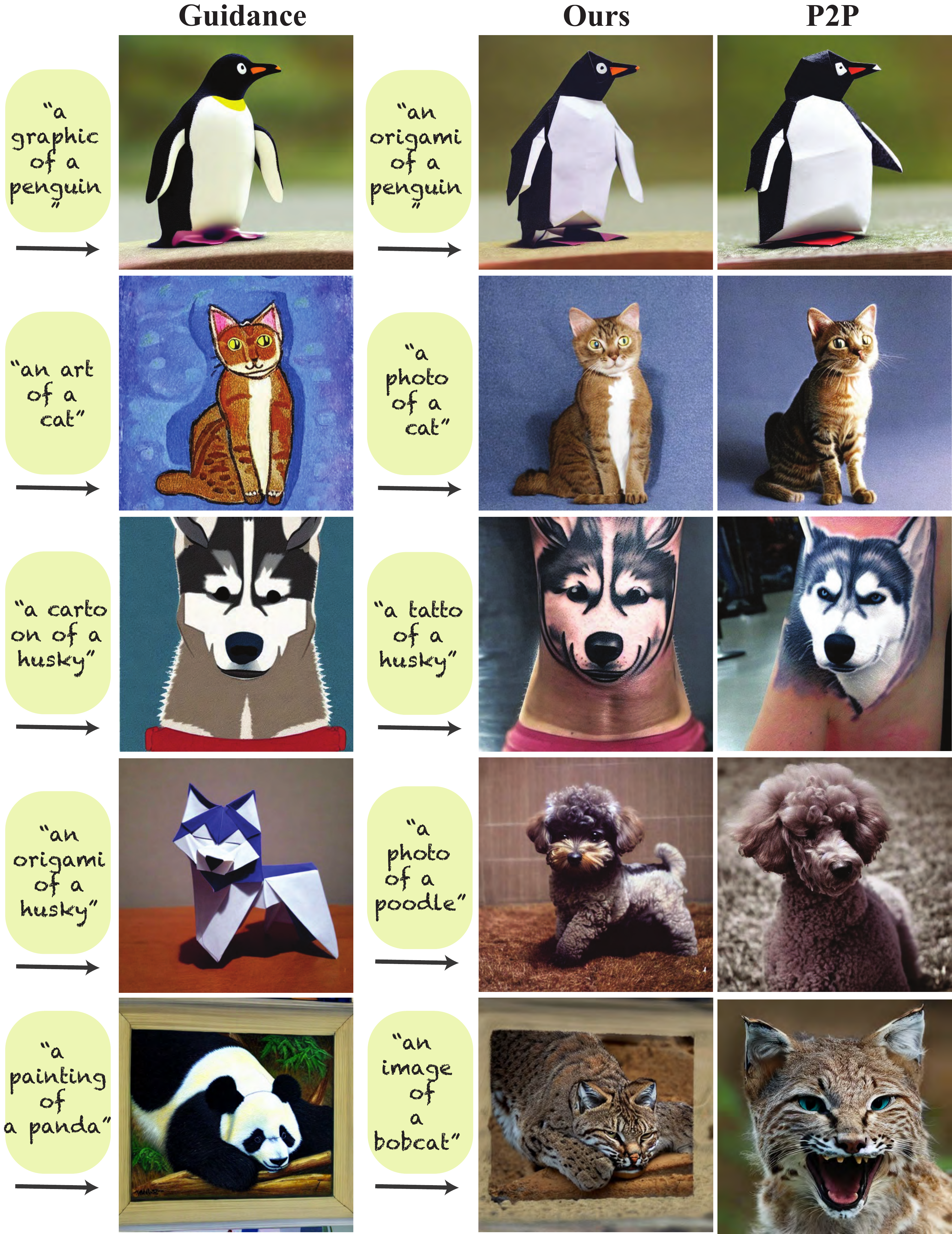}
    \caption{\emph{Comparison to P2P on generated \emph{ImageNet-R-TI2I} benchmark.}  While P2P results demonstrate high fidelity to the  the target text, there are noticeable deviation from the guidance structure, especially in cases of multiple word swaps (last two rows). Across all examples, our results adhere to  the target edit while preserving the guidance  scene layout and object pose. }
    \label{fig:qualitative_fakeImnetR}\afterfigure
\end{figure}
\paragraph{Extended comparison to P2P~\protect\cite{hertz2022prompt_to_prompt}}
To factor out the effect of DDIM inversion, we expand our comparison to P2P on \emph{generated} guidance images. Specifically, we created a \emph{generated-ImageNet-R-TI2I} benchmark by using text prompts expressing the same object classes and renditions described in \cref{section:appendix-imnetr}.  

As seen in \cref{fig:qualitative_fakeImnetR}, both our method and P2P comply with the target text. However, P2P often results in large deviations from the guidance structure, especially in cases where multiple prompts edits are applied (last two rows in \cref{fig:qualitative_fakeImnetR}). Our method demonstrates fine-grained structure preservation across all these examples, while successfully translating multiple traits (e.g., both category and style). 
These properties are strongly evident by \cref{fig:quant-comp}, where our method results in a significantly lower self-similarity distance, even compared to P2P with injecting cross-attention at all timesteps ($t=1000$).

\begin{figure}[t!]
    \centering
    \includegraphics[width=\linewidth]{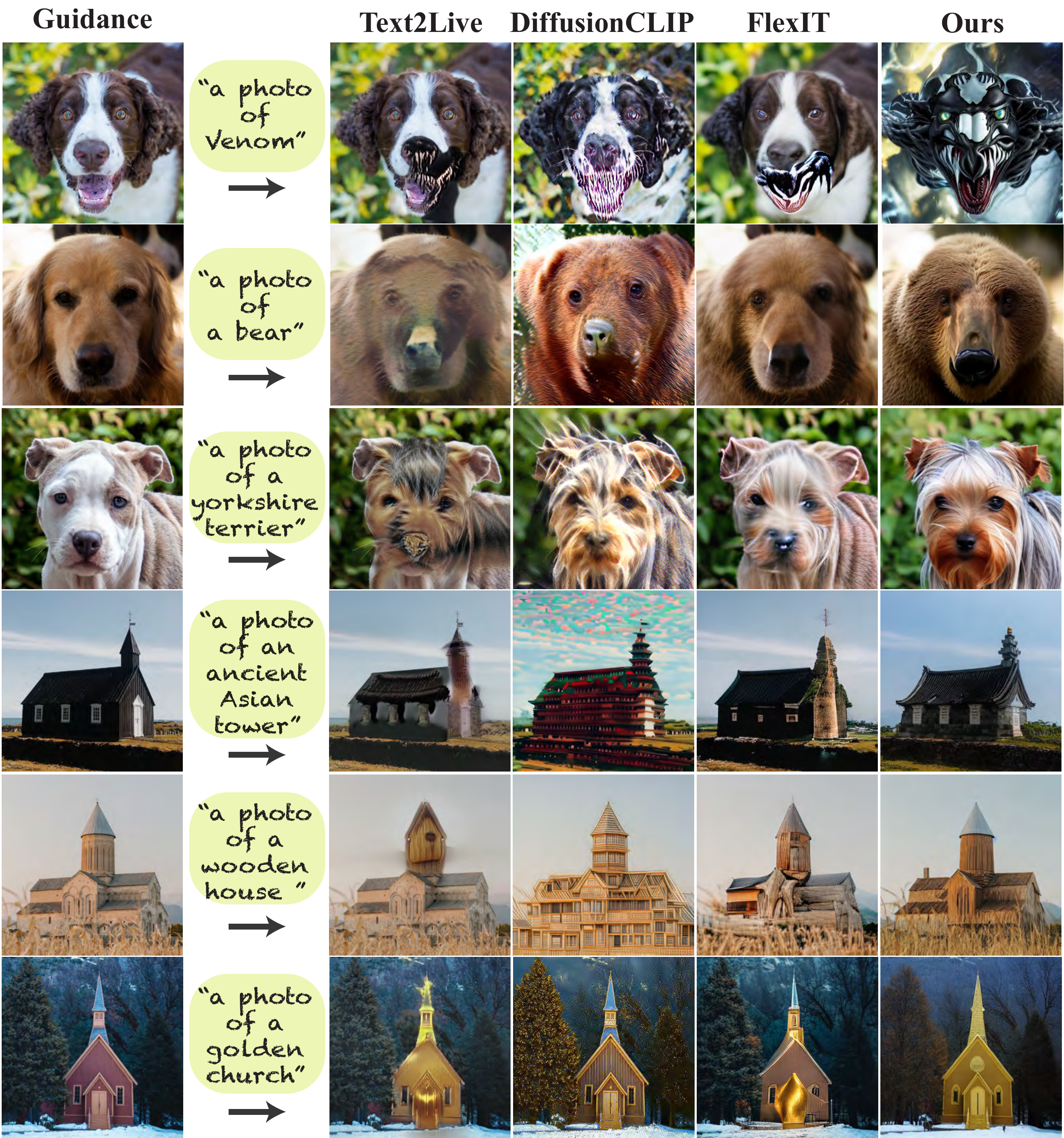}
    \caption{Qualitative comparisons to additional baselines: Text2LIVE \cite{bar2022text2live}, DiffusionCLIP \cite{kim2022diffusionclip}, FlexIT \cite{flexit2021}. These methods fail to deviate from the structure for matching the target prompt, or create undesired artifacts.\label{fig:additional}
    }\afterfigure
\end{figure}

\paragraph{Additional baselines.} Fig.~\ref{fig:additional} shows qualitative comparisons with: (i) Text2LIVE \cite{bar2022text2live}, (ii) DiffusionCLIP \cite{kim2022diffusionclip}, and (iii) FlexIT \cite{flexit2021}. These methods either fail to deviate from the guidance image or result in noticeable visual artifacts. 
Since Text2LIVE is designed for layered textural editing, thus it can only ``paint'' over the guidance image and cannot apply any structural changes necessary to convey the target edit (e.g. dog to venom, church to Asian tower). Moreover, Text2LIVE does not leverage a strong generative prior, hence often results in low visual quality. FlexIT often fails to deviate from the guidance content, which may be caused to their regularization that encourages the guidance and output images to match in LPIPS sense. We also note that DiffusionCLIP requires fine-tuning an unconditional diffusion model for each target edit on a set of 30+ images from a single domain (e.g. dog faces, churches).

\subsection{Ablation}
We ablate our key design choices by evaluating our performance for the following cases: (i) w/o spatial features injection (w/o features), (ii) w/o self-attention injection. The metrics are reported in Fig.~\ref{fig:quant-comp}(a) and a representative example is shown in Fig.~\ref{fig:ablation}.  The results demonstrate that both features and self-attention are critical for structure preservation -- the features provide a semantic association between the original and translated content, while self-attention is essential for maintaining this association and capturing finer structural information. 
Further ablations can be found in \cref{sec:appendix-ablations} and \cref{table:sup-imnet-r}.

\begin{figure}[h!]
\centering  
\includegraphics[width=1\linewidth]{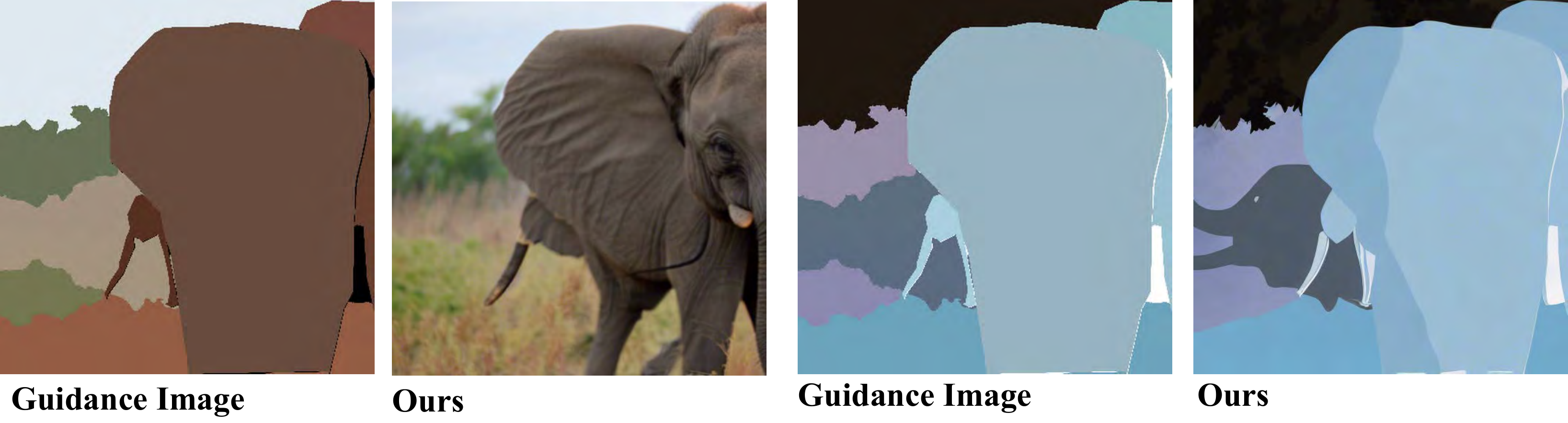}
\caption{\emph{Limitations.} Our method fails when there is no semantic association between the guidance content and the target text. Thus, it does not perform well on solid segmentation masks with arbitrary colors.} \afterfigure
\label{fig:limitation}
\end{figure}

\section{Discussion and Conclusion}
We presented a new framework for diverse text-guided image-to-image translation, founded on new insights about the internal representation of a pre-trained text-to-image diffusion model. Our method, based on simple manipulation of features, outperforms existing baselines, achieving a significantly better balance between preserving the guidance layout and deviating from its appearance. As for limitations, our method relies on the semantic association between the original and translated content in the diffusion feature space. Thus, it does not work well on detailed label segmentation masks where regions are colored arbitrarily (Fig.~\ref{fig:limitation}). In addition, we are relying on DDIM inversion, which we found to work well in most of our examples. Nevertheless, we observed that for textureless ``minimal'' images, DDIM may occasionally result in a latent that encodes dominant low-frequency appearance information, in which case some appearance information would leak into our results.
We believe that our work demonstrates the yet unrealized potential of the rich and powerful feature space spanned by pre-trained text-to-image diffusion models. We hope it will motivate future research in this direction.

\paragraph{Acknowledgments:} 
We thank Omer Bar-Tal for his insightful comments and discussion. 
This project received funding from the Israeli Science Foundation (grant 2303/20), the Carolito Stiftung, 
 and the NVIDIA Applied Research Accelerator Program.
Dr. Bagon is a Robin Chemers Neustein AI Fellow.

\vfill

{\small
\bibliographystyle{ieee_fullname}
\bibliography{egbib}
}
\appendix
\newpage

\twocolumn[{
\renewcommand\twocolumn[1][]{#1}
\maketitle
\centering
\vspace*{-0.73cm}
}]

\section{Ablations}\label{sec:appendix-ablations}
\begin{table*}
\centering
    \renewcommand{\tabcolsep}{4pt}
\begin{tabular}{c | c  c  c | c  c  c | c  c  c} 
& \multicolumn{3}{c|}{Wild-TI2I Real} & \multicolumn{3}{c|}{Wild-TI2I Generated} & \multicolumn{3}{c}{ImageNet-R-TI2I} \\
\cline{2-10}
\noalign{\vskip\doublerulesep
         \vskip-\arrayrulewidth}
\cline{2-10}
& Self-Sim $\downarrow$ & CLIP $\uparrow$ & LPIPS $\uparrow$ & Self-Sim $\downarrow $& CLIP $\uparrow$ & LPIPS $\uparrow$  & Self-Sim $\downarrow$ & CLIP $\uparrow$ & LPIPS $\uparrow$   \\

 \hline
w/ encoder-feat-7
& 0.058 & 0.280 &  0.527 & 0.035 & 0.264 & 0.453 & 0.05 & 0.273 & 0.458 \\
\hline
 w/o neg. prompt & 0.052 & 0.281 & 0.490 & 0.033 & 0.275 & 0.441  & 0.048 & 0.274 & 0.451 \\
 \hline
 w/o feat. & 0.090 & 0.288 & 0.584  & 0.084 & 0.297 & 0.633 & 0.076 & 0.281 & 0.534 \\ 
 \hline
 w/o self-attn. & 0.097 & 0.286 & 0.597  & 0.090 & 0.295 & 0.657  & 0.089 & 0.278 & 0.564 \\
  \hline
\textbf{Our method}
& 0.058 & 0.282 & 0.521  & 0.048 & 0.289 & 0.542  & 0.051 & 0.275 & 0.462 \\
 \hline
\end{tabular}
\caption{\textbf{Quantitative evaluation on WILD-Real benchmark}. We evaluate the distance in DINO-ViT self-similarity for structure preservation, CLIP score for target text faithfulness and LPIPS distance for deviation from the guidance image. We ablate the features injection, self-attention injection, negative prompting, and additional feature injection in the encoder blocks. We report these scores on our three text guided I2I benchmarks. The configuration reported in the main paper (encoder features + self-attention injection and negative prompting) is the best balance between the three metrics across the datasets.
\label{table:sup-imnet-r}}
\end{table*}

\subsection{Negative-prompting.}
\label{sec:appendix-np}

\begin{figure}
    \centering
    \includegraphics[width=\linewidth]{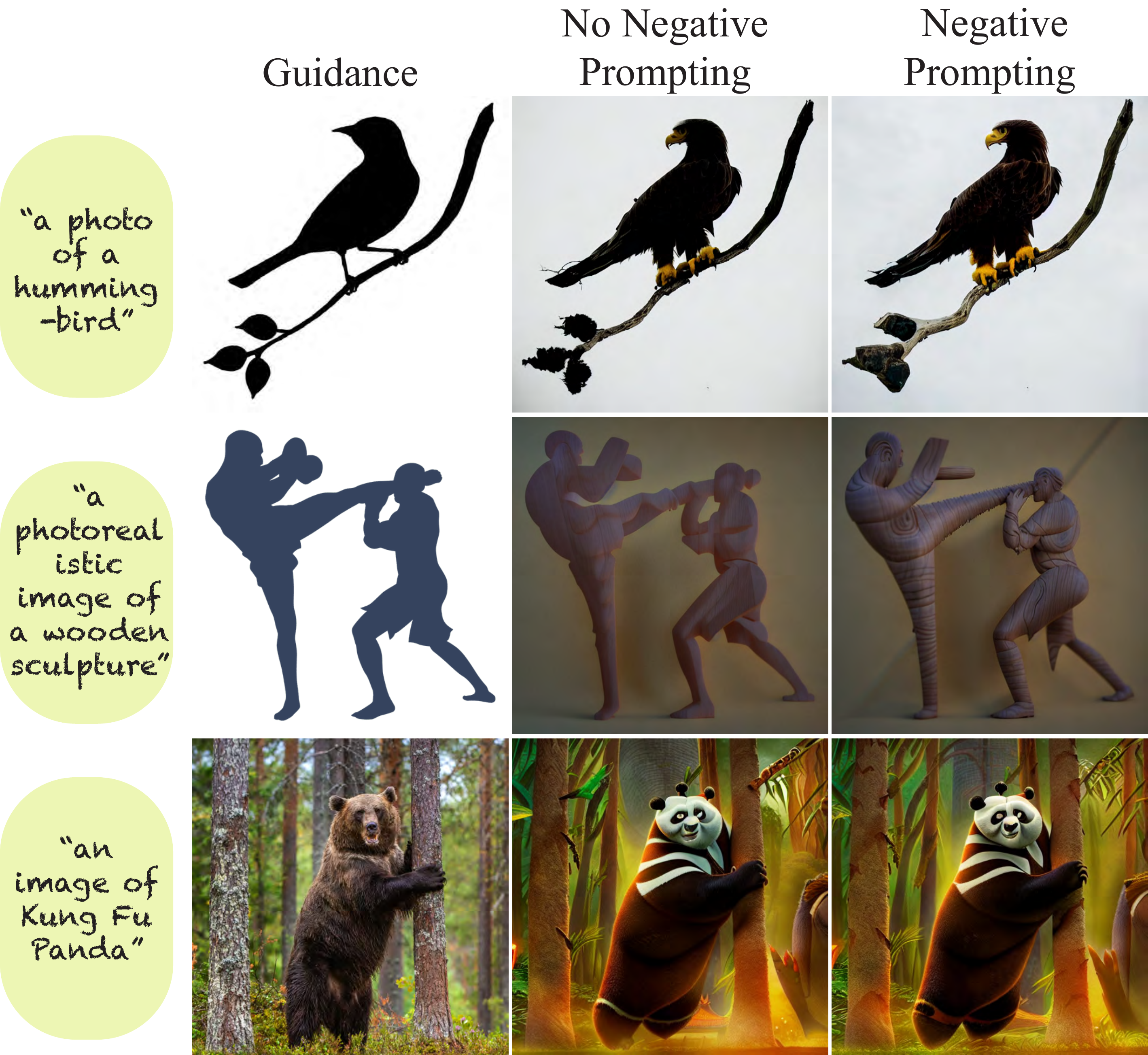}
    \caption{\emph{Qualitative ablation of negative prompting.} The effect of negative prompting is most meaningful on textureless guidance images. In the case of realistic images (row 3) it has a minor effect.}
    \label{fig:np-ablation}
\end{figure}

We qualitatively and quantitatively ablate the effect of negative prompting (see \cref{sec:method} of the main paper).    \cref{table:sup-imnet-r} compares our metrics w/ and w/o negative prompting (bottom row and second row), using our \emph{Wild-TI2I} and \emph{ImageNet-R-TI2I} benchmarks. The results indicate that the usage of negative prompting (bottom row) leads to slightly larger deviation from the guidance image (higher LPIPS distance between $I^G$ and $I^*$), while introducing only a minor reduction in structure preservation. Sample results of this ablation are shown in \cref{fig:np-ablation}, where we can also notice that negative prompting has a larger effect for  ``primitive images'', i.e, simple ``textureless" images such as silhouettes (top two rows) than natural guidance images. 

\subsection{Injected features.}
\label{sec:appendix-pca}
\begin{figure*}
    \centering
    \includegraphics[width=0.9\linewidth]{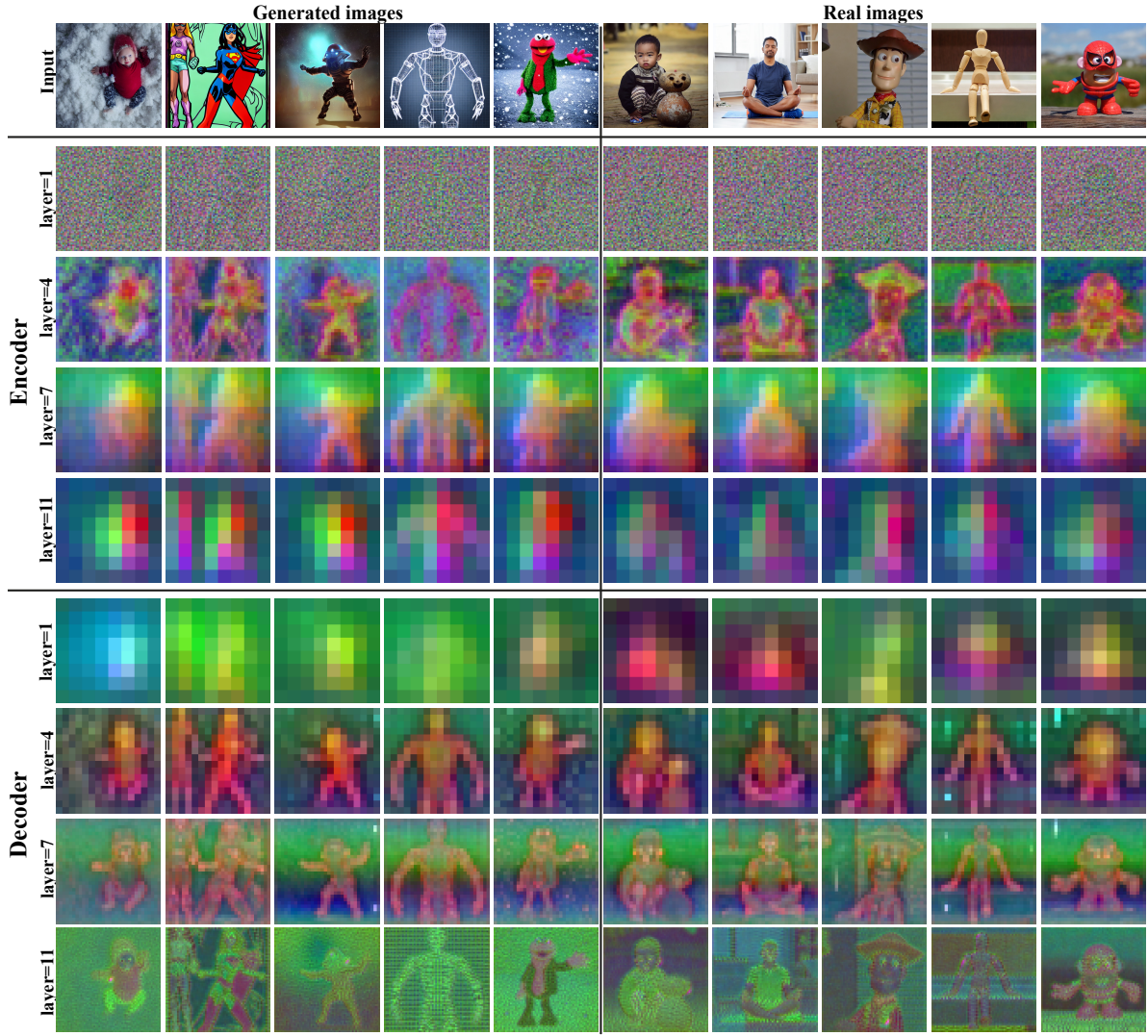}
    \caption{\emph{Visualizing diffusion features for both encoder and decoder.} 
    Extending the visualization of \cref{fig:pca1} in the main paper to include features from encoder blocks of the U-Net at time $t\!=\!540$ (top part).
    \label{fig:pca-full}}
\end{figure*}

Our method injects features to the \emph{decoder} block, in a specific layer which we observed to capture localized semantic information. To complete our analysis, we extend our PCA feature visualization to include both the decoder and encoder features. As seen in Fig.~\ref{fig:pca-full}, the encoder  resemble a mirrored trend to the decoder: the encoder features start with high frequency noise (layer 1), which is gradually transformed into cleaner features that depict lower-frequency content throughout the layers. Nevertheless, localized semantic information is overall less apparent in the encoder's features. To numerically evaluate this, we consider a modified version of our method where the encoder features from layer 7, which resemble some semantic information,  are additionally injected. As seen in \cref{table:sup-imnet-r}, this combination results in worse CLIP score in all data-sets and smaller LPIPS deviation from the guidance image on most sets (first row).

\begin{figure}
    \centering
    \includegraphics[width=.8\linewidth]{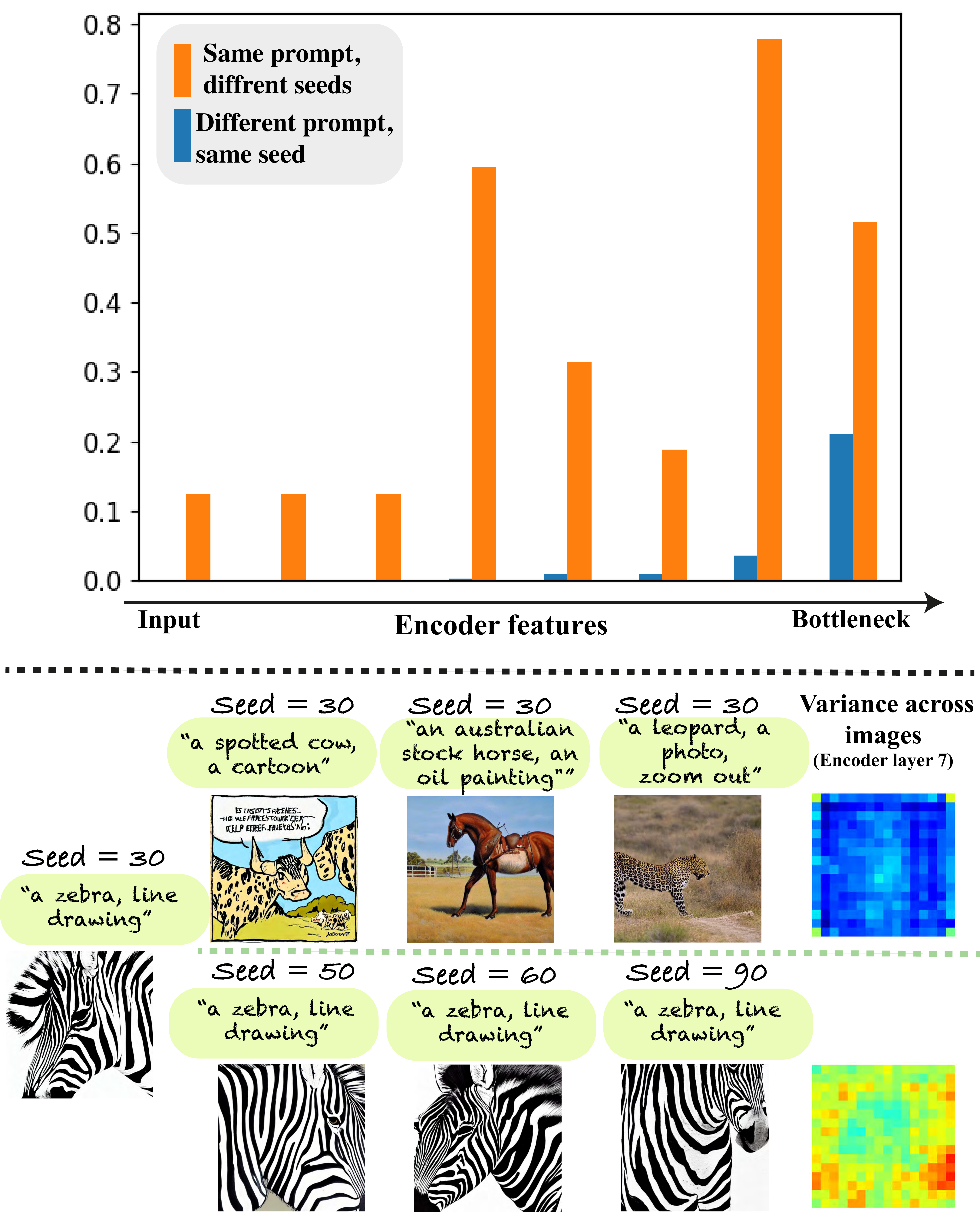}
    \caption{\emph{Measuring features' variance: different prompts vs. different seeds.} We consider 10 different seeds and 10 different prompts, and generate 100 images using all combinations.   We extract  features from the encoder (which are skipped to the decoder) for all generated images at  $t=T$,
    and compute the variance over features originated from: (i)~the same prompt across different seeds, and (ii)~the same seed across different prompts. The estimated variance under each of these settings is plotted in orange and blue bars, respectively. We observe that although the generated images using (ii) do not exhibit shared visual properties, their features are correlated (low variance). In contrast, images generated using (i) are more visually similar, yet their features are significantly less correlated.} \afterfigure
    \label{fig:var}
\end{figure}

\section{Initial noise $\bs{x}_T$ and spatial features} 

\label{subsec:appendix-init-noise}

We observed that in order for our method to work, the initial noise used to generate the translated image $\bs{x}^*_T$ has to  match the initial guidance noise $\bs{x}^G_T$.  
Since we inject features into the decoder from the very first step of the backward process, this dependency on the random seed can only be explained by the encoder features at $t\!=\!T$, denoted by $\bs{f}_t^{e_l}$. Recall that these features depend on both $\bs{x}^*_T$ and the target prompt $P$. This raises the question: why $\bs{f}_T^{e_l}$ originated from $\bs{x}^*_T = \bs{x}^G_T$ and an arbitrary text prompt $P$, allow our method to work?  We hypothesize that in $t\!=\!T$ the target prompt has little effect on the encoder features $\bs{f}_T^{e_l}$, thus the injected decoder features $\bs{f}^4_T$ comply with the encoder features.  In contrast, changing the seed results in a mismatch between $\bs{f}^4_T$ and $\bs{f}_T^{e_l}$.  This may be surprising since  images generated from the same seed under different text prompts may dramatically differ from one another (see Fig.~\ref{fig:var} bottom).

To validate this hypothesis, we performed an analysis that shows that  features formed from the same  $\bs{x}_T$ under arbitrary prompts $\{P_i\}$ are significantly more correlated than those generated under the same prompt $P$ with different seeds $\{\bs{x}^i_T\}$. 
Specifically, we used 10 different prompts and 10 different seeds to generate 100 images, using all combinations. We considered the images generated under:  (i)~the same prompt across different seeds, and (ii)~the same seed across different prompts. Total of 20 sets of 10 images each.  We then measured the  variance between the feature maps within each of these sets.
In \cref{fig:var} (top), we report these  variances (averaged across spatial location) as a function of the encoder layer $l$. As seen, changing the initial seed, for any fixed prompt, results in significantly higher variance across features for all layers $l$ compared to fixing the seed and changing only the prompt. These findings validate our hypothesis and support our method's dependency on the initial seed.

\section{Implementation Details}
\label{sec:appendix-implementation}

We use Stable Diffusion as our pre-trained text-to-image model;  we use the \emph{StableDiffusion-v-1-4} checkpoint provided via \href{https://huggingface.co/CompVis/stable-diffusion-v-1-4-original}{official HuggingFace webpage}.

In all of our experiments, we use DDIM deterministic sampling with 50  steps. In the case of real guidance images, we perform deterministic DDIM inversion with 1000 forward steps and then perform deterministic DDIM sampling with 1000 backward steps. Our translation results are performed with 50 sampling steps, thus we extract features only at these steps. We set our default  injection thresholds to: $\tau_{A} = 25$, $\tau_{f}=40$ out of the 50 sampling steps; for primitive guidance image, we found that $\tau_{A} = \tau_{f} = 25$ to work better.

 During translation, we set the classifier-free guidance scale for real and generated guidance images to 15.0 and 7.5, respectively.
The use of negative prompting is controlled via a hyperparameter $\alpha$ used to interpolate between the predicted noise $\epsilon_\theta(N)$  and the predicted noise $\epsilon_\theta(\empty)$, as described in \cref{sec:method}. We set 
an initial value $\alpha_0$ in the first sampling step, and then gradually decrease it. For real guidance images, we set $\alpha_0 = 1.0$ and use a linear scheduler $\alpha(t) = \alpha_0 - t$. For generated guidance images, we set $\alpha_0 = 0.75$. We use the linear scheduler for realistic generated images, and for generated images that are primitive/textureless, we use an exponential scheduler $\alpha(t) = e^{-6 \cdot t}$. 

For running the competitors, we use their official implementations: \href{https://github.com/google/prompt-to-prompt}{Prompt-to-Prompt}, \href{https://github.com/anon294384/diffuseit}{DiffuseIT}, \href{https://colab.research.google.com/drive/1E8QHZ3BbkF6hzk0rRKzhfkySmYf_BZaE?usp=sharing}{DiffusionCLIP}, \href{https://github.com/omerbt/Text2LIVE}{Text2LIVE}, \href{https://github.com/facebookresearch/SemanticImageTranslation}{FlexIT}. For running SDEdit on StableDiffusion, we use the implementation available in the \href{https://github.com/CompVis/stable-diffusion/blob/main/scripts/img2img.py}{Stable Diffusion official repo}. For running VQGAN-CLIP, we used \href{https://github.com/nerdyrodent/VQGAN-CLIP}{the publicly available repo}.

\section{Benchmarks}
\label{sec:appendix-benchmarks}
\subsection{ImageNet-R-TI2I benchmark.}
\label{section:appendix-imnetr}
To test our method on a wide range of guidance images, we turn to Image-Net-R~[15], a dataset that contains various renditions of 200 classes from Image-Net. 
We manually select 10 classes: ``\pr{Castle}", ``\pr{Cat}", ``\pr{Goldfish}", ``\pr{Hummingbird}", ``\pr{Husky}", ``\pr{Jeep}", ``\pr{Panda}", ``\pr{Penguin}", ``\pr{Pizza}", ``\pr{Violin}".
To avoid low-quality images, we manually selected 3 images per class, totaling 30 guidance images.

Additionally, we automatically created 5 target prompts per image.
All the prompts share the same template: ``$\ll$rendition$\gg$ of a $\ll$class$\gg$", e.g. ``a painting of a jeep".
$\ll$rendition$\gg$ is one of the existing renditions in the real ImageNet-R data-set: ``\pr{an art}", ``\pr{a cartoon}", ``\pr{a graphic}", ``\pr{a deviantart}", ``\pr{a painting}", ``\pr{a sketch}", ``\pr{a graffiti}", ``\pr{an embroidery}", ``\pr{an origami}", ``\pr{a pattern}", ``\pr{a sculpture}", ``\pr{a tattoo}", ``\pr{a toy}", ``\pr{a video-game}", ``\pr{a photo}", ``\pr{an image}".
For two (out of five) target prompts per image, we changed the correct $\ll$class$\gg$ to another object class randomly sampled from 5 related classes (to avoid completely unreasonable translations such as penguin $\rightarrow$ jeep).

Overall, our \emph{ImageNet-R-TI2I} benchmark contains 150 image-text pairs: 30 guidance images with 5 target prompts each.

\subsection{Wild TI2I benchmark.}
We collected a diverse dataset of 148 text-image pairs, containing different object classes (people, animals, food, landscapes) in different renditions (realistic images, drawings, solid masks, sketches and illustrations) with different levels of semantic details. 53\% of the examples consists of real guidance images that we gathered from the Web, and the rest are generated from text.

We will publicly release our benchmarks and code for academic use.

\end{document}